\documentclass{article}

\usepackage[preprint]{neurips_2024}

\usepackage[utf8]{inputenc} 
\usepackage[T1]{fontenc}    
\usepackage{hyperref}       
\usepackage{url}            
\usepackage{booktabs}       
\usepackage{amsfonts}       
\usepackage{amsmath}

\usepackage{nicefrac}       
\usepackage{microtype}      
\usepackage{xcolor}         
\usepackage{pifont}
\usepackage{dsfont}
\usepackage{hyperref}
\usepackage{booktabs}
\usepackage{bbm}
\usepackage{adjustbox}
\newcommand{\cmark}{\ding{51}} 
\usepackage{algorithm}
\usepackage{algorithmic}
\usepackage{xcolor, colortbl}
\usepackage{multirow}
\usepackage{graphicx}

\title{CLARIFID: Improving Radiology Report Generation by Reinforcing Clinically Accurate Impressions and Enforcing Detailed Findings}

\author{%
  Kyeongkyu Lee\thanks{These authors contributed equally.}, Seonghwan Yoon\footnotemark[1], Hongki Lim\thanks{Corresponding author.} \\
  Department of Electrical and Computer Engineering\\
  Inha University\\
  Incheon 22212, South Korea \\
  \texttt{rudrb1231@inha.edu, ysh1210@inha.edu, hklim@inha.ac.kr} \\
}

\begin{document}

\maketitle

\begin{abstract}
Automatic generation of radiology reports has the potential to alleviate radiologists’ significant workload, yet current methods struggle to deliver clinically reliable conclusions. In particular, most prior approaches focus on producing fluent text without effectively ensuring the factual correctness of the reports and often rely on single-view images, limiting diagnostic comprehensiveness. We propose CLARIFID, a novel framework that directly optimizes diagnostic correctness by mirroring the two-step workflow of experts. Specifically, CLARIFID (1) learns the logical flow from Findings to Impression through section-aware pretraining, (2) is fine-tuned with Proximal Policy Optimization in which the CheXbert F1 score of the Impression section serves as the reward, (3) employs controlled decoding that completes ``Findings'' before synthesizing the ``Impression'', and (4) fuses multiple chest X-ray views via a vision-transformer-based multi-view encoder. During inference, we apply a next-token forcing strategy followed by report-level re-ranking, ensuring that the model first produces a comprehensive “Findings” section before synthesizing the ``Impression'' and thereby preserving coherent clinical reasoning. Experimental results on the MIMIC-CXR dataset demonstrate that our method achieves superior clinical efficacy and outperforms existing baselines on clinical efficacy scores. The source code is available at \url{https://github.com/IncheonYSH/CLARIFID}.
\end{abstract}

\section{Introduction}\label{}
Chest X-rays (CXRs) are the most widely used medical imaging modality for diagnosing conditions affecting the heart, lungs, mediastinum, and bone. However, converting these images into accurate and comprehensive diagnostic narratives requires considerable radiologist time and expertise, creating a bottleneck in clinical workflows. Radiology Report Generation (RRG) aims to alleviate this burden by automatically translating medical images into structured, clinically meaningful narratives.

{
Over the past few years, radiology report generation (RRG) has progressed along two axes that matter directly for clinical utility. 
The first of these concerns representation. Models strengthen long‐range clinical context and image–text alignment with memory or hierarchy enhanced encoder–decoders~\cite{chen-emnlp-2020-r2gen, chen-etal-2021-cross-modal}.
To curb hallucinations and bias toward frequent normalities, many systems inject structured priors: medical knowledge graphs, entity-relation schemas, or contrastive objectives that pull abnormal cues closer in representation space~\cite{Zhang_Wang_Xu_Yu_Yuille_Xu_2020, jain2021radgraphextractingclinicalentities, 9578840, kiut, dynamicgraphcontrastive, wang2022multi, CoFe}.
A complementary line explicitly grounds textual spans to anatomical evidence via detector-aligned attention or region-phrase alignment, improving localization and explainability~\cite{RGRG, 10587279}.
Concurrently, LLM-based approaches leverage lightweight visual adapters and diagnosis-driven prompts to better capture radiology-specific style and higher-level reasoning without retraining massive backbones~\cite{10.1609/aaai.v38i3.28038, wang2023r2gengptradiologyreportgeneration, chang2024bootstrapping}.
Multi-view CXR processing has also advanced; however, practical simplifications—such as fixing view configurations to rigid single/dual-view setups or reshaping reports to fit single-image paradigms—can inadvertently limit access to inter-view nuances and clinically informative comparison cues, leaving complementary projections under-leveraged in practice~\cite{MLRG,10.1609/aaai.v39i8.32890}. 

The other axis concerns the training objective. Beyond cross-entropy on image–report pairs, reinforcement learning (RL) has been explored to align training with evaluation signals; however, rewards are often based on text-overlap metrics (BLEU/ROUGE/METEOR) that only partially capture clinical correctness~\cite{qin-song-2022-reinforced}.
To better reflect radiologists’ judgments, semantically informed objectives reward entity consistency or graph coherence—e.g. RadGraph based signals to reduce hallucinations and strengthen factual grounding~\cite{miura-etal-2021-improving, jain2021radgraphextractingclinicalentities, delbrouck-etal-2022-improving}.
Recent work further considers multi-objective preference formulations to accommodate heterogeneous clinical criteria across conditions and report sections~\cite{xiao2024radiologyreportgenerationmultiobjective}.
In parallel, clinically grounded automatic labelers (e.g., CheXbert) have enabled scalable, label-space-aware assessment of clinical efficacy, motivating objectives that move beyond surface similarity toward decision-relevant accuracy~\cite{smit-etal-2020-combining}.
Despite these advances, three gaps remain: reliance on supervision that requires labeling pipelines beyond standard image–report pairs, proxy-similarity optimization, and under-emphasis of the Impression—gaps that motivate our design.
}

\begin{figure}[t]
	\centering
	\includegraphics[width=0.7\columnwidth]{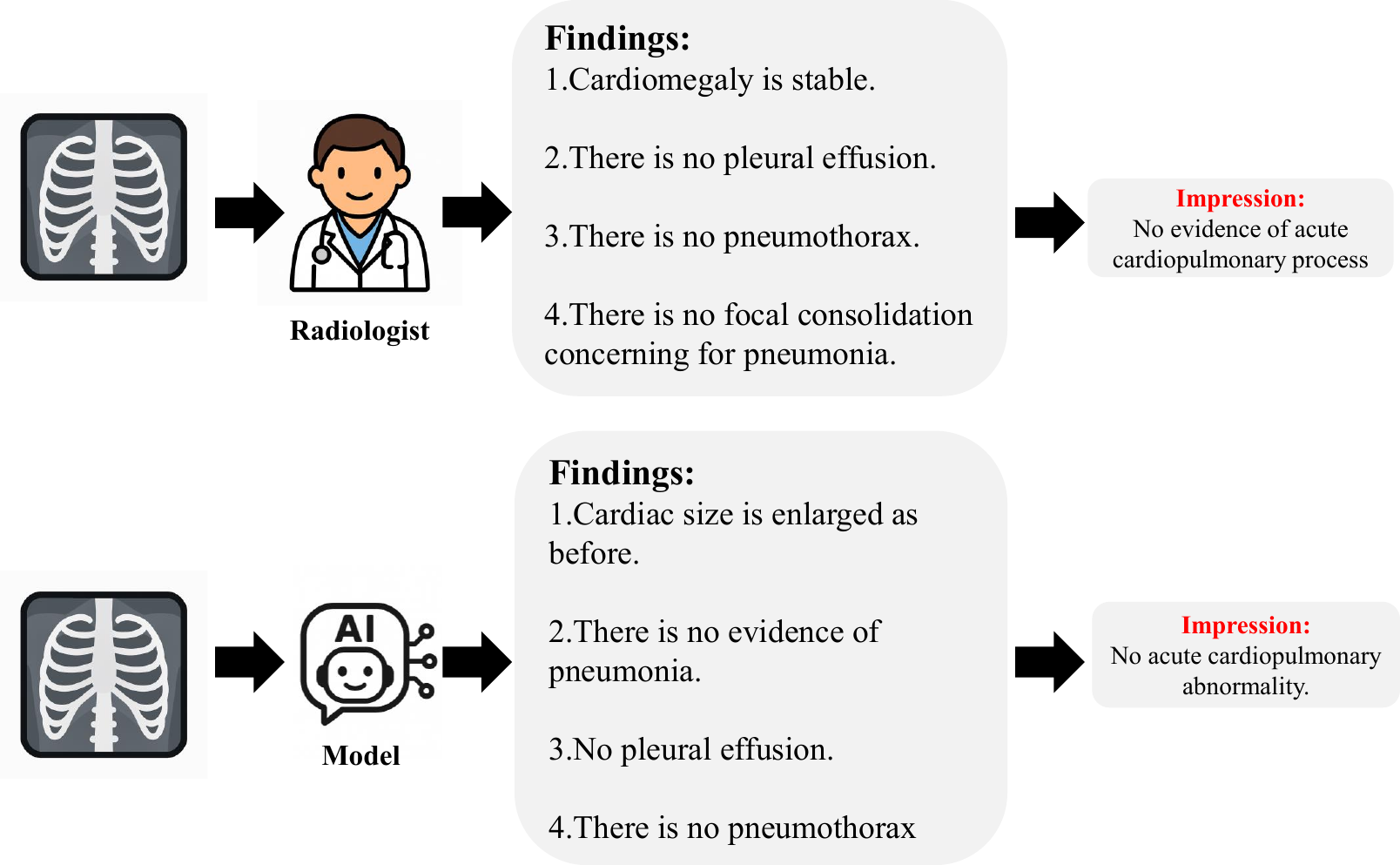}
	\caption{Radiologist and CLARIFID follow the same clinical workflow, recording the Findings first and then deriving the Impression. This demonstrates the model's alignment with expert practice.}
	
	\label{fig:enter-label}
\end{figure}

The most widely used datasets, MIMIC-CXR~\cite{johnson2019mimiccxrjpglargepubliclyavailable} and IU X-ray~\cite{demner2016preparing}, both provide reports in which the Findings and Impression sections are clearly separated. Notably, in MIMIC-CXR, 82.4\% of diagnostic labels are derived from the Impression section~\cite{johnson2019mimiccxrjpglargepubliclyavailable}, highlighting its fundamental clinical importance. Yet most prior work in radiology report generation has prioritized generating detailed Findings with less attention to the diagnostic synthesis required for the Impression, creating a significant gap between research and clinical practice.

We reframe radiology report generation as a structured clinical reasoning task that mirrors the radiologist's workflow: first documenting detailed observations (``Findings''), then synthesizing them into diagnostic conclusions (``Impression'') (see Figure~\ref{fig:enter-label}). The Impression serves as the ``answer'' that directly informs clinical decision-making, while Findings provide supporting evidence and context. Analogous to recent advances in language model reasoning---which show that outcome-based optimization improves overall performance and can correlate with higher-quality intermediate steps~\cite{orm2021, zelikman2024quietstar, R1, shen2025vlmr1stablegeneralizabler1style, r1like2}---we hypothesize that optimizing for clinical accuracy of the Impression section will encourage higher-quality Findings documentation through implicit learning of the relationship between observations and diagnoses.

Building on this insight, we introduce CLARIFID (\textbf{CL}inically \textbf{A}ccu\textbf{R}ate \textbf{I}mpressions and detailed \textbf{FI}n\textbf{D}ings), a novel framework that enhances medical factuality through reinforcement learning directly optimized for clinical accuracy of the Impression section. Unlike previous methods that rely on cross-entropy loss or optimize for surface-level text similarity, our approach aligns the training objective with diagnostic correctness by using the CheXbert-based F1 score as a reward signal. By explicitly rewarding the model for generating clinically faithful impressions, we implicitly encourage the generation of more relevant and supportive findings, creating a virtuous cycle of improved clinical reasoning.

To further support high-quality report generation during inference, we introduce a controlled decoding strategy that prevents premature transition to the Impression section. We delay ``Impression'' tokens with next‑token forcing until the ``Findings'' section reaches a preset length and then apply our section‑aware temperature schedule, using a higher temperature for Findings to surface detail and a lower one for Impression to maintain focus.

Furthermore, radiologists routinely review multiple projections from a single CXR study, but prior studies are often inaccessible~\cite{Ge2012}, and many existing RRG frameworks either overlook this reality~\cite{chen-emnlp-2020-r2gen,chen-etal-2021-cross-modal, wang2023r2gengptradiologyreportgeneration,10.1609/aaai.v38i3.28038} or adopt overly complex architectures~\cite{MLRG}. To address this limitation, we introduce a flexible multi‑view fusion module capable of integrating any number of CXR images in one study.

Our main contributions are as follows:
\begin{itemize}  
        \item We develop section-aware pretraining that encodes the flow from Findings to Impression via sentence-level control tokens, laying the foundation for structured reasoning.

    \item We propose an Impression-centric reinforcement learning framework, directly optimizing for clinical accuracy.
	
	\item We introduce a reasoning‑aware decoding strategy specialized for radiology report generation that reflects both clinical intent and report structure.
	
	\item We adapt a multi‑view fusion architecture to process variable‑length X‑ray inputs from a single study.
\end{itemize}

\section{Related work}
\label{related_work}
\subsection{Reinforcement learning for RRG}
{
We view reinforcement learning (RL) in radiology report generation (RRG) as a framework for optimizing 
non-differentiable, decision-oriented objectives that maximum-likelihood estimation (MLE) does not directly capture. 
A policy generates a candidate report, a learned or rule-based clinical reward evaluates its utility, and the policy is updated via policy-gradient or actor–critic methods. Preference-based alignment (e.g., RLHF/RLAIF/DPO) extends this 
mechanism by learning from human or AI preferences~\cite{ouyang2022traininglanguagemodelsfollow,pmlr-v235-lee24t,rafailov2024directpreferenceoptimizationlanguage}; 
however, in clinical settings such feedback is costly, privacy-restricted, and often \emph{section-agnostic}, making it difficult to enforce 
Impression-centric objectives.

Within RRG, early RL mainly optimized proxy text-overlap rewards (BLEU/ROUGE/METEOR)~\cite{qin-song-2022-reinforced}, 
and subsequent work employed semantically grounded signals such as entity or graph consistency (e.g., RadGraph) to curb hallucinations and improve factual grounding~\cite{miura-etal-2021-improving,jain2021radgraphextractingclinicalentities,delbrouck-etal-2022-improving}. 
In contrast, we adopt a \emph{section-dependent} objective that targets \emph{Impression-level} clinical correctness via a CheXbert-derived signal~\cite{smit-etal-2020-combining}, 
while relying only on standard image–report pairs (no auxiliary labels or detectors). Optimization details are provided in Section~{\ref{sec:method}}.
}

\subsection{Test-time scaling strategies}
Test-time scaling methods enhance model performance by leveraging additional computational resources during inference. Instead of modifying model parameters or extending the training process, these methods enrich the inference stage itself. Broadly, they fall into two categories: parallel and sequential approaches.

Parallel methods generate multiple candidate outputs and select the best one using techniques such as majority voting or best-of-N sampling with a reward model~\cite{DBLP:conf/iclr/0002WSLCNCZ23, brown2024largelanguagemonkeysscaling, irvine2023rewardingchatbotsrealworldengagement,  NEURIPS2024_51173cf3, song2024goodbadgreedyevaluation}. These approaches improve output quality through sampling, without altering model weights.

Sequential approaches, in contrast, iteratively refine a single reasoning path. Starting with an initial response, the model revisits and improves its output based on predefined criteria such as logical consistency, task-specific constraints, or self-generated feedback. Techniques like self-refine~\cite{NEURIPS2023_91edff07}, reflexion~\cite{shinn2023reflexionlanguageagentsverbal}, and decomposition-based self-correction~\cite{palmeira-ferraz-etal-2024-llm, Li_2022}, as well as chain-of-thought prompting~\cite{wei2023chainofthoughtpromptingelicitsreasoning}, exemplify this approach. These methods deepen reasoning without requiring multiple samples, offering a compute-efficient path to improved performance.
Budget forcing~\cite{muennighoff2025s1simpletesttimescaling} and adaptive injection decoding~\cite{jin2025wellthinkingenhancingllm} further illustrate sequential scaling. These methods regulate inference-time computation by either forcibly terminating generation or prolonging reasoning through repeated insertion of control tokens, particularly when the model attempts to conclude prematurely. Unlike majority voting, whose performance often saturates with more samples, token forcing strategies can benefit from increased reasoning depth.

Together, these strategies demonstrate the flexibility and effectiveness of test-time scaling in enhancing LLM performance across diverse generation tasks.

\begin{figure*}[t]
	\centering
	\includegraphics[width=\textwidth]{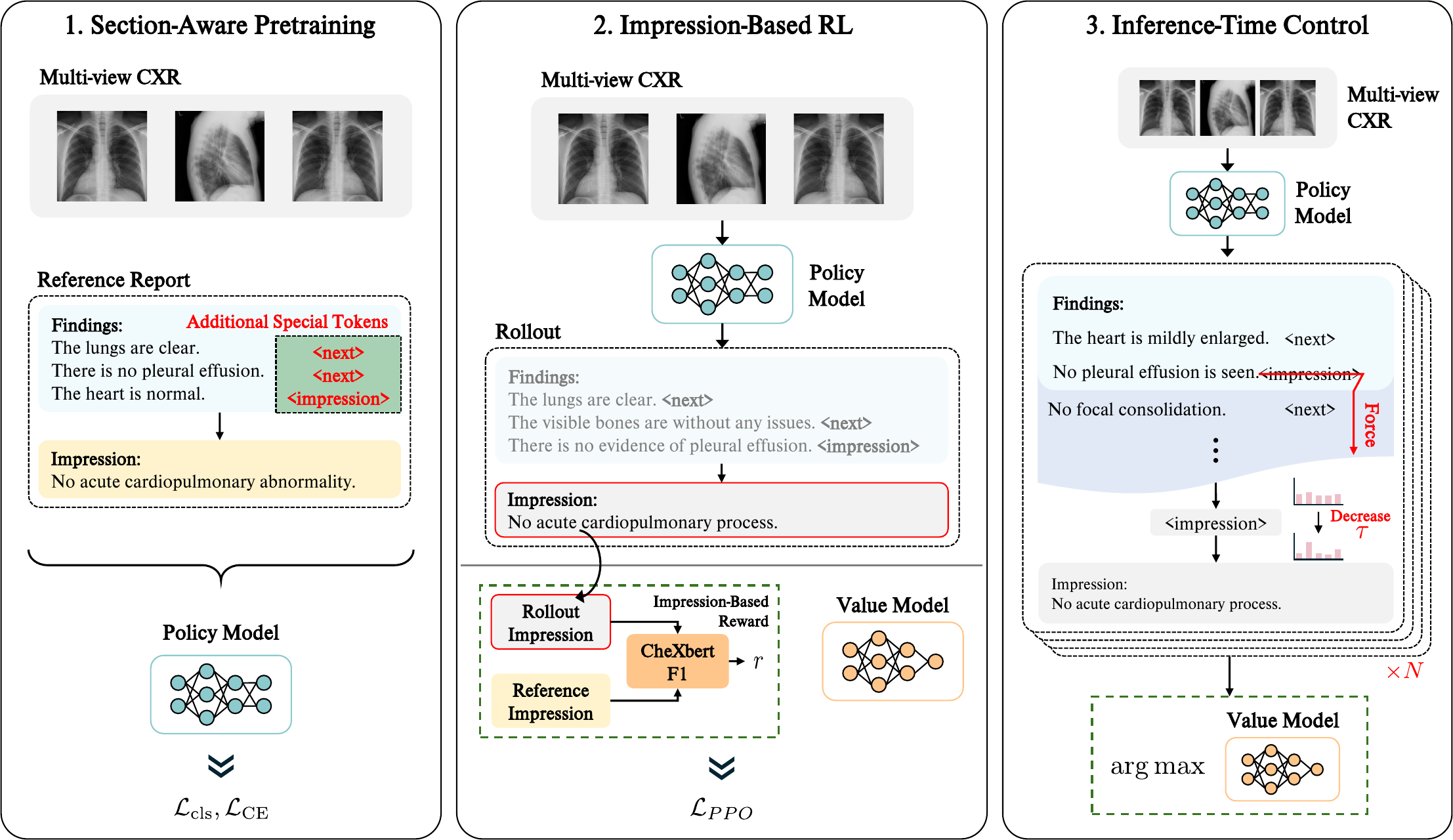}
	\caption{Overview of the CLARIFID framework.  
		(1) Section-aware pretraining flags each Findings sentence with \texttt{<next>} and the Impression with \texttt{<impression>}, teaching the report’s two-step structure.  
		(2) After a supervised warm-up, the model is fine-tuned by PPO, using CheXbert-F1 on the Impression as reward.  
		(3) At inference, \texttt{<impression>} is withheld until at least $k$ \texttt{<next>} tokens have been generated; the model then samples $N$ complete candidate reports and outputs the top-scoring report. {Refer to Figure \ref{fig:model-arch} for details on the architectures of the policy and value models.}}
	\label{fig:overview}
\end{figure*}

\section{Method}\label{sec:method}
Our CLARIFID framework integrates multi-view image encoding, structured pretraining, reinforcement learning, and controlled decoding to generate clinically faithful radiology reports. Figure~\ref{fig:overview} illustrates the complete framework. In this section, we elaborate on each component in detail.

\subsection{Model architecture}
\begin{figure}[t]
  \centering
  \includegraphics[width=\linewidth]{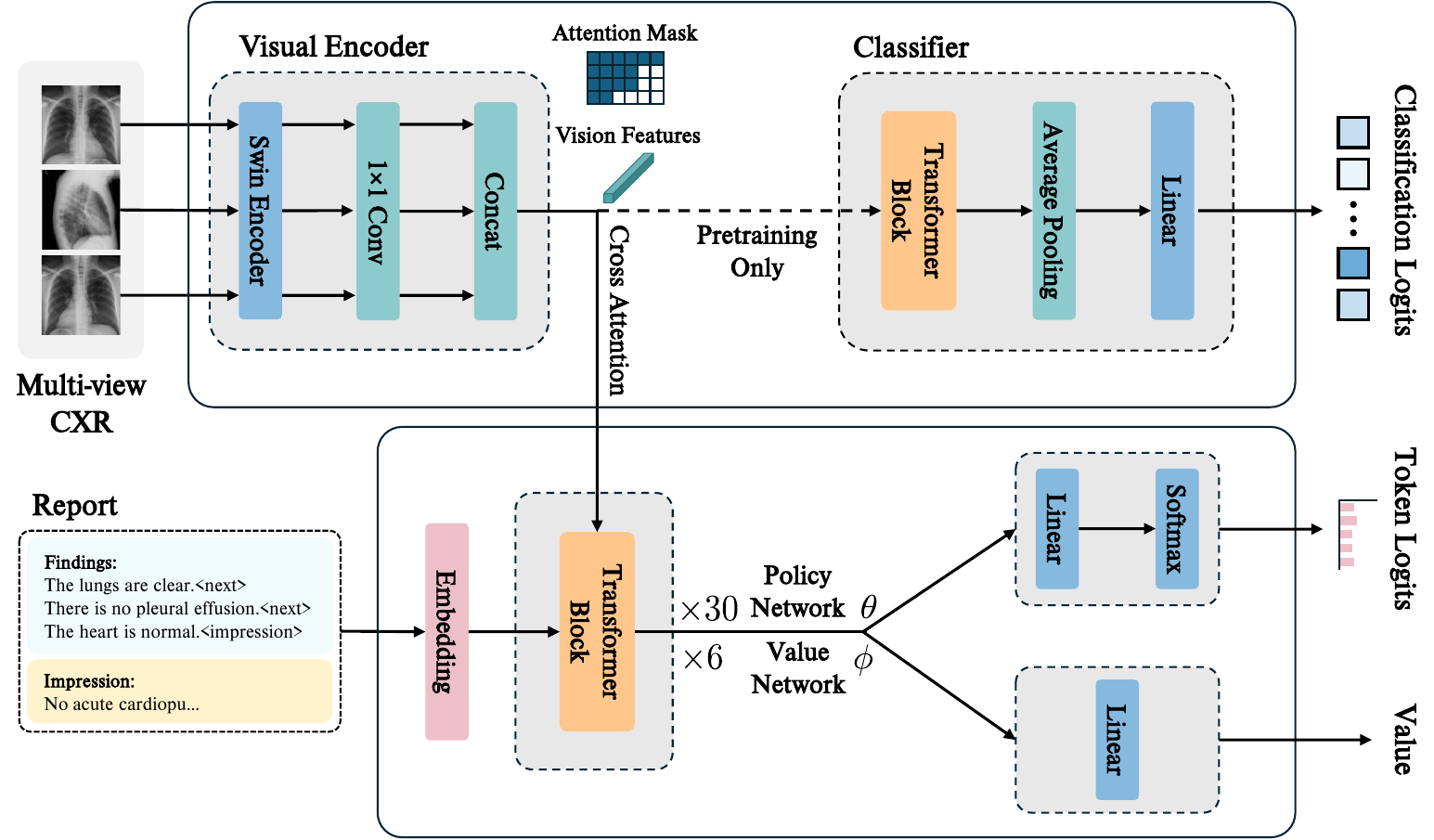}
  \caption{Model architecture schematic diagram summarizing modules and data flow.}
  \label{fig:model-arch}
\end{figure}

Both the policy and value networks are based on the GPT-2 architecture~\cite{radford2019language}.  
Visual information is supplied by a SwinV2~\cite{swinv2}-based multi-image encoder that fuses all available X-ray views of a study.  
Conditioned on these fused features, the policy network generates reports, while the value network estimates expected rewards for reinforcement learning.  
The two networks use different depth configurations to balance performance and computational efficiency. {A compact schematic with the module-level connections is provided in Figure~\ref{fig:model-arch}.}

\subsubsection{Multi-image encoder}
Radiologists typically examine several X-ray views (e.g., PA, AP, LAT). To mirror this workflow, our model processes a variable number of images per study.
For a batch of $B$ studies, where study $i$ has $K_i$ images, each image $I_{i,j}$ is encoded by a Swin Transformer encoder $E$. Let the last-stage feature map have spatial size $H_f \times W_f$ and channel $D_{\mathrm{img}}$. We apply a $1{\times}1$ convolution $W_{\mathrm{proj}}:\mathbb{R}^{D_{\mathrm{img}}}\!\to\!\mathbb{R}^{D}$ to match the model hidden size $D$, then flatten to $N{=}H_fW_f$ patch tokens, yielding $X_{i,j}\in\mathbb{R}^{N\times D}$.

For study $i$, we concatenate its $K_i$ image-token sequences
\(
X^{\mathrm{concat}}_i=[X_{i,1};\,X_{i,2};\,\dots;\,X_{i,K_i}] \in \mathbb{R}^{(K_iN)\times D},
\)
so $M_i{=}K_iN$ tokens are available per study. Packing a batch and padding to $M_{\max}{=}\max_i M_i$ produces padded image feature
\(
\widehat{\mathbf{S}}\in\mathbb{R}^{B\times M_{\max}\times D}, \)
and attention mask
\(
\mathbf{A}\in\{0,1\}^{B\times M_{\max}},
\)
where $A_{i,m}{=}1$ denotes a valid visual token and $0$ denotes padding. We feed $(\widehat{\mathbf{S}}, \mathbf{A})$ directly to the text decoder’s cross-attention as keys/values and the encoder mask, respectively.

\subsubsection{Policy network architecture}
The policy network employs a decoder with 30 transformer layers, hidden size $D=768$, and 12 attention heads. It processes the padded image representations $\widehat{\mathbf{S}} \in \mathbb{R}^{B \times M_{\text{max}} \times D}$ and attention mask $\mathbf{A} \in \{0,1\}^{B \times M_{max}}$ through cross-attention at each transformer layer:
\begin{align}
	h_l^{self} &= \text{SelfAttn}(h_{l-1}) \\
	h_l^{cross} &= \text{CrossAttn}(h_l^{self}, \widehat{\mathbf{S}}, \text{mask}=\mathbf{A}) \\
	h_l &= \text{FFN}(h_l^{cross})
\end{align}

In the cross-attention operation, we use learnable projections $W_Q,W_K,W_V,W_O\in\mathbb{R}^{D\times D}$. Queries, keys, and values are
\[
\mathcal Q \;=\; h_l^{\text{self}} W_Q \in \mathbb{R}^{B\times T\times D},\quad
\mathcal K \;=\; \widehat{\mathbf{S}} W_K \in \mathbb{R}^{B\times M_{\max}\times D},\quad
\mathcal V \;=\; \widehat{\mathbf{S}} W_V \in \mathbb{R}^{B\times M_{\max}\times D}.
\]
From the binary padding mask $\mathbf{A}$ we form a mask $\mathcal{M}\in\mathbb{R}^{B\times 1\times M_{\max}}$ with
$\mathcal{M}_{i,1,m}{=}-\infty$ if $A_{i,m}{=}0$ and $0$ otherwise (broadcast across $T$). The attention and output are
\[
\alpha \;=\; \mathrm{softmax}\!\left(\frac{\mathcal Q \mathcal K^\top}{\sqrt{D}} + \mathcal{M}\right)\in\mathbb{R}^{B\times T\times M_{\max}},\qquad
\widetilde{H}_l \;=\; \alpha \mathcal V \in\mathbb{R}^{B\times T\times D},\qquad
h_l^{\text{cross}} \;=\; \widetilde{H}_l W_O.
\]
The attention mask $\mathbf{A}$ ensures that for each study $i$, only the first $K_i \cdot N$ entries in row $i$ (representing valid visual tokens) influence the generation.
This architecture enables clinically accurate reports conditioned on all available views, with the policy network outputting a distribution $P_\theta(y_t|y_{<t}, \widehat{\mathbf{S}}, \mathbf{A})$ at each step, where $h_l\in\mathbb{R}^{B\times T\times D}$(sequence length $T$) is the hidden state of layer $l$, $y_t$ is the target token at step $t$, $y_{<t}$ denotes the prefix tokens up to $t{-}1$, and $\theta$ collects all policy network parameters.

\subsubsection{Value network architecture}
The value network employs a decoder with 6 transformer layers, hidden size $D=768$, and 12 attention heads. It adopts the same cross-attention interface and inputs $(\widehat{\mathbf{S}}, \mathbf{A})$ as the policy network, but replaces the language-modeling head with a position-wise linear value head. Let $h^{\text{final}}\in\mathbb{R}^{B\times T\times D}$ denote the final hidden states of the value decoder. We evaluate the value function $V_\phi$ at the autoregressive state
\(s_{i,t}=(y_{i,<t},\,\widehat{\mathbf{S}}_i,\,\mathbf{A}_i)\).
The value head \(f_\phi\) is an affine map applied independently at each token position with parameters \(\mathbf{w}_\phi\in\mathbb{R}^{D}\) and \(b_\phi\in\mathbb{R}\).
\[
\mathbf{v} = f_\phi(h^{\text{final}}) = h^{\text{final}} \,\mathbf{w}_\phi + b_\phi \in \mathbb{R}^{B \times T}, \qquad
V_\phi(s_t) = \mathbf{v}_{t}.
\]
Here $\phi$ collects all parameters of the value network.

\subsection{Section-aware pretraining}
We first warm up the image encoder and the lightweight aggregator on a 14-condition multi-label classification task.
For a batch of $B$ studies with variable numbers of images, we reuse the padded visual tokens $\widehat{\mathbf{S}}\in\mathbb{R}^{B\times M_{\max}\times D}$ and the binary mask $\mathbf{A}\in\{0,1\}^{B\times M_{\max}}$.
A lightweight Transformer encoder $f_{\mathrm{agg}}$ aggregates the tokens and a masked average pooling produces a study-level representation:
\[
\mathbf{H} = f_{\mathrm{agg}}(\widehat{\mathbf{S}},\,\mathbf{A}) \in \mathbb{R}^{B\times M_{\max}\times D}, \qquad
\mathbf{h}_i = \frac{1}{Z_i}\sum_{m=1}^{M_{\max}} A_{i,m}\,\mathbf{H}_{i,m}, \quad
Z_i=\sum_{m=1}^{M_{\max}}A_{i,m}.
\]
Here, $m$ indexes the padded visual-token sequence (concatenated across all images), and $A_{i,m}{=}1$ if the $m$-th token of study $i$ is valid.

After masked average pooling, a linear classification head maps each $\mathbf{h}_i\in\mathbb{R}^{D}$ to 14 logits with parameters $W_{\mathrm{cls}}\in\mathbb{R}^{14\times D}$ and $\mathbf{b}_{\mathrm{cls}}\in\mathbb{R}^{14}$.
\begin{align}
\mathbf{z}_i = W_{\mathrm{cls}}\mathbf{h}_i + \mathbf{b}_{\mathrm{cls}} \in \mathbb{R}^{14}, \qquad
z_{i,c} = [\mathbf{z}_i]_c.
\label{eq:classification_logits}
\end{align}
We then convert CheXbert~\cite{smit-etal-2020-combining} condition labels to binary targets, 
where \texttt{positive} is mapped to $1$ and \texttt{blank}, \texttt{negative}, and \texttt{uncertain} 
are mapped to $0$. Let $\tilde{g}_{i,c} \in \{\texttt{blank},\texttt{positive},\texttt{negative},\texttt{uncertain}\}$ 
denote the original label. Define
\begin{align}
g_{i,c} = \mathbbm{1}[\tilde{g}_{i,c} = \texttt{positive}] =
\begin{cases}
1, & \tilde{g}_{i,c} = \texttt{positive},\\
0, & \tilde{g}_{i,c} \in \{\texttt{blank},\texttt{negative},\texttt{uncertain}\}
\end{cases}
\label{eq:binary_label}
\end{align}

We optimize a per-condition weighted binary cross-entropy with weights $w_c$ estimated on the training split to mitigate class imbalance, where $w_c = \frac{n_c^-}{\,n_c^+}$ with $n_c^+$ and $n_c^-$ denoting the numbers of positive and negative instances for condition $c$ on the training split, and with $\sigma(\cdot)$ denoting the logistic function:

\begin{align}
\mathcal{L}_{\mathrm{cls}}
= - \frac{1}{B}\sum_{i=1}^{B}\sum_{c=1}^{14}
\Bigl[
w_c\, g_{i,c}\,\log\sigma(z_{i,c})
+ (1-g_{i,c})\,\log\!\bigl(1-\sigma(z_{i,c})\bigr)
\Bigr]
\label{eq:weighted_bce}
\end{align}

After image encoder warm-up, to explicitly represent the two-stage structure of radiology reports, consisting of detailed observations (Findings) followed by a diagnostic summary (Impression), we then adopt a section-aware pretraining scheme that uses control tokens to mark the report’s logical structure at the sentence level.
During training, we insert \texttt{<next>} tokens between consecutive findings sentences and prepend a special \texttt{<impression>} token before the impression section. This structured approach helps the model learn section-specific style and content, distinguishing between the detailed, observational language of Findings and the concise, conclusive language of Impressions. It also establishes clear section boundaries, preventing content bleeding between sections that could compromise clinical accuracy. Additionally, these special tokens create natural breakpoints for controlled decoding during inference, enabling our next-token forcing strategy.

The decoder is trained with a cross-entropy objective, jointly optimized with the classification loss $\mathcal{L}_{\mathrm{cls}}$, conditioned on the multi-image encoder’s outputs:
\begin{align}
\mathcal{L}_{\mathrm{CE}} &= -\sum_{t=1}^T \log P_\theta\bigl(y_t \mid y_{<t}, \widehat{\mathbf{S}}, \mathbf{A}\bigr)\label{eq:ce_loss}\\
    \mathcal{L}_{\mathrm{pre}} &= \mathcal{L}_{\mathrm{cls}} + \mathcal{L}_{\mathrm{CE}}\label{eq:pre_loss}
\end{align}
Where $\mathcal{L}_{\mathrm{CE}}$ is the cross-entropy loss, $\mathcal{L}_{\mathrm{cls}}$ is the weighted binary cross-entropy loss for image classification, and $\mathcal{L}_{\mathrm{pre}}$ is the joint objective for section-aware pretraining. This supervised pretraining stage establishes a foundation of radiological knowledge and reporting conventions before the reinforcement learning phase.

\begin{algorithm}[t]
\centering
\begin{minipage}{0.9\columnwidth}
\caption{CLARIFID Training}
\label{alg:clarifid_training}
\begin{algorithmic}[1]
\REQUIRE Dataset $\mathcal{D}$, generator $\pi_{\theta}$, generator image encoder $E_{\theta}$, value model $V_{\phi}$,
         pretraining cycles $C_{\mathrm{pre}}$, image-only cycles $C_{\text{img}}$,
         PPO iterations $C_{\mathrm{ppo}}$, Value function loss weight $\lambda_V$

\STATE \textbf{Stage I: Section-aware supervised pretraining}
\STATE {\textbf{Init:} Add the special tokens \texttt{\textless next\textgreater} and \texttt{\textless impression\textgreater} to the tokenizer.}
\FOR{$\mathrm{iter} \gets 1$ to $C_{\mathrm{pre}}$}
    \STATE \textit{Repeat over mini-batches.}
    \STATE Encode images to get $(\widehat{\mathbf{S}},
    \mathbf{A})$.
    \STATE Compute 14 logits using~\eqref{eq:classification_logits}.
    \STATE Binarize ground-truth label with~\eqref{eq:binary_label}
    \IF{$\mathrm{iter} \le C_{\text{img}}$}
        \STATE \textbf{\# image encoder warm-up}
        \STATE Update $E_{\theta}$ using the weighted BCE loss in~\eqref{eq:weighted_bce}.
    \ELSE
        \STATE \textbf{\# section-aware LM pretraining}
        \STATE Build targets: insert \texttt{<next>} in Findings, prepend \texttt{<impression>}.
        \STATE Update $\pi_{\theta}$ with joint objective~\eqref{eq:pre_loss} conditioned on $(\widehat{\mathbf{S}}, \mathbf{A})$.
    \ENDIF
\ENDFOR

\STATE \textbf{Stage II: Impression-based reinforcement learning}
\STATE $\pi_{\text{ref}} \gets \pi_\theta$.
\FOR{$\mathrm{iter} \gets 1$ to $C_{\mathrm{ppo}}$}
    \STATE Collect a set of trajectories $\mathcal{T}$ from interactions using current policy $\pi_\theta$
     \FOR{each trajectory $\tau \in \mathcal{T}$}
        \STATE $r \gets \mathrm{F1}\!\big(\mathrm{CheXbert}(\mathrm{Impression}_{\text{gen}}), \mathrm{CheXbert}(\mathrm{Impression}_{\text{gt}})\big)$
        \STATE Compute augmented rewards $R'$ using ~\eqref{eq:kl_penalty}
        \STATE Estimate advantages $\hat{A}$ with \eqref{eq:advantage} -- \eqref{eq:advantage_whitening}
    \ENDFOR
     \FOR{$\mathrm{epoch} \gets 1$ to $E_{\mathrm{ppo}}$}
        \FOR{each mini-batch of data from $\mathcal{T}$}
            \STATE Update $\theta, \phi$ by optimizing $\mathcal{L}_{\text{PPO}} + \lambda_V\mathcal{L}_{{V}}$ from~\eqref{eq:ppo_objective} -- \eqref{eq:total_loss}
        \ENDFOR
    \ENDFOR
\ENDFOR
\STATE \textbf{return} $\pi_\theta$
\end{algorithmic}
\end{minipage}
\end{algorithm}

\subsection{Impression-based reinforcement learning}
After supervised pretraining, we employ Proximal Policy Optimization (PPO) to directly optimize the clinical utility of generated reports. Unlike previous approaches that rely on text-similarity metrics or entity-level rewards, we specifically target the Impression section's clinical accuracy using CheXbert-based F1 scores as the primary reward signal. We expect that optimizing Impression-level clinical accuracy as the reinforcement signal will encourage the model to refine the Findings section as an intermediate reasoning step, thereby improving its factual alignment with the final diagnostic conclusions.

\subsubsection{PPO framework}
PPO~\cite{schulman2017proximalpolicyoptimizationalgorithms} is a policy gradient method that provides stable updates by constraining the policy change between iterations. For our report generation task, we conceptualize the problem in terms of states, actions, policies, and rewards. The states ($s_t$) consist of the current partial report and visual features, while actions ($a_t$) represent the selection of the next token in the sequence. The policy ($\pi_\theta$) defines the probability distribution over possible next tokens.

\paragraph{Rollout and reward structure.}
At each PPO iteration, we perform $N$ stochastic rollouts per input using temperature sampling, which increases on-policy exploration and provides a richer set of trajectories without additional gradient steps. The reward formulation is token-level, consisting of (i) a per-token KL penalty against a reference policy $\pi_{\text{ref}}$ to preserve fluency and (ii) a terminal clinical reward from the CheXbert-based F1 score on the Impression section.

\paragraph{Policy regularization.}
A major challenge in reinforcement learning fine-tuning is preventing policy collapse, where the model sacrifices linguistic quality and coherence to maximize a sparse, terminal reward. To address this, we introduce KL-divergence regularization, which provides a dense reward signal to guide the policy at every generation step. This per-token KL term acts as a shaping reward, penalizing the current policy $\pi_\theta$ if it strays too far from the fluent distribution of the supervised pre-trained reference policy $\pi_{\text{ref}}$. The final reward is augmented as follows:
\begin{align}
R' = R - \beta \cdot \mathrm{KL}(\pi_\theta \,\|\, \pi_{\text{ref}})
\label{eq:kl_penalty}
\end{align}
Here, $\beta$ is a fixed coefficient that balances the trade-off between maximizing the clinical accuracy reward ($R$) and maintaining the linguistic quality inherited from the reference model. This provides a stable regularization effect throughout the fine-tuning process.

\paragraph{Generalized advantage estimation.}
A key component of PPO training is the estimation of the advantage function, which quantifies how much better (or worse) an action $a_t$ is compared to the expected value of state $s_t$. Directly using Monte Carlo returns provides unbiased estimates but suffers from very high variance, while bootstrap-based temporal-difference methods exhibit low variance but are biased.
To balance this trade-off, we adopt Generalized Advantage Estimation (GAE)~\cite{schulman2018highdimensionalcontinuouscontrolusing}, which interpolates between these two extremes using the parameter $\lambda \in [0,1]$. 
To estimate advantages, we use Generalized Advantage Estimation (GAE) with discount factor $\gamma=1$ and trace-decay parameter $\lambda=0.95$:

\begin{align}
	\delta_t &= r_t + \gamma V_\phi(s_{t+1}) - V_\phi(s_t),
    \qquad
    \hat{A}_t = \sum_{l=0}^{T-t-1} (\gamma\lambda)^l \delta_{t+l}
\label{eq:advantage}
\end{align}

We then apply advantage whitening~\cite{schulman2017proximalpolicyoptimizationalgorithms, OAIbaseline, huang2024nimplementationdetailsrlhf} over non-padded positions:
\begin{align}
\tilde{A}_t = \frac{\hat{A}_t - \mu_A}{\sigma_A}
\label{eq:advantage_whitening}
\end{align}

which reduces variance and improves numerical stability.

\paragraph{Objective function.}
Finally, the policy parameters are updated by maximizing the clipped PPO~\cite{schulman2017proximalpolicyoptimizationalgorithms} surrogate objective. The standard policy gradient objective can be written in terms of the probability ratio 
$r_t(\theta) = \frac{\pi_\theta(a_t|s_t)}{\pi_{\theta_{\text{old}}}(a_t|s_t)}$, 
which measures how much the new policy deviates from the old policy for the same action.
 If $r_t(\theta) > 1$, the action becomes more likely under the new policy, and if $r_t(\theta) < 1$, it becomes less likely.

Directly optimizing $r_t(\theta)\tilde{A}_t$ can lead to excessively large updates when $r_t(\theta)$ deviates far from 1, causing training instability. To mitigate this, PPO introduces a clipping mechanism:
\begin{align}
\mathcal{L}_{\mathrm{PPO}}
= \mathbb{E}_t \!\left[
  \min \Big(
    r_t(\theta)\,\tilde{A}_t,\;
    \operatorname{clip}\!\big(r_t(\theta),\,1-\epsilon,\,1+\epsilon\big)\,\tilde{A}_t
  \Big)
\right]
\label{eq:ppo_objective}
\end{align}
where $\epsilon$ limits how much the new policy is allowed to diverge from the old one. The clipping ensures that when $r_t(\theta)$ goes outside the range $[1-\epsilon, 1+\epsilon]$, the update is truncated, preventing the policy from moving too far in a single step.

Intuitively, this surrogate objective preserves the benefits of large advantage updates when the policy is close to the old one, while penalizing overly aggressive updates that might destabilize learning. Together with GAE and KL regularization, the clipped objective forms the core of PPO’s stability, balancing exploration with conservative policy improvement.

\paragraph{Value function training.}
 To ensure stable advantage estimates, the value function $V_{\phi}$ is trained alongside the policy. It is updated by minimizing a clipped mean squared error loss, which prevents the value network from undergoing drastic changes that could destabilize policy updates. The target for this update is empirical return  $R_t=\hat{A}_t + V_{\mathrm{old}}(s_t)$, where $V_{\mathrm{old}}(s_t)$ represents the value estimate from before the update step. Similar to the policy objective, the new value prediction $V_{\phi}(S_t)$ is clipped to stay within an $\epsilon$-ball of the old value $V_{\phi_{old}}(S_t)$.
 
\begin{align}
\mathcal{L}_{{V}}
=\frac{1}{2}\,\mathbb{E}_{t\in\mathcal{M}}
\Big[
\max\!\Big( \big(V_\phi(s_t)-R_t\big)^2,\;
\big(\mathrm{clip}(V_\phi(s_t),\,V_{\mathrm{old}}(s_t){\pm}\epsilon\big)-R_t)^2 \Big)
\Big]
\label{eq:value_loss}
\end{align}
At each update, we optimize a joint objective combining the clipped PPO surrogate and the value loss:
\begin{align}
\mathcal{L}_{\text{total}} \;=\; \mathcal{L}_{\text{PPO}} \;+\; \lambda_V\,\mathcal{L}_{{V}}
\label{eq:total_loss}
\end{align}
Where $\lambda_V$ is the value loss weight.

\subsubsection{Reward formulation}
After generating a complete report, we extract the Impression section and evaluate it using the CheXbert model, which assigns labels to 14 common radiological conditions. The reward is calculated as the micro-averaged F1 score between the CheXbert-extracted labels from the generated impression and the ground truth.
To stabilize optimization, we apply advantage whitening after estimating returns, reducing variance across trajectories while preserving the natural scale of the CheXbert-F1 reward.
\[
r = \text{F1}(\text{CheXbert}(\text{Impression}_{gen}), \text{CheXbert}(\text{Impression}_{gt}))
\]
Algorithm~\ref{alg:clarifid_training} summarizes the entire training pipeline, from section-aware pretraining to impression-based reinforcement learning.

\begin{algorithm}[t]
\centering
\begin{minipage}{0.9\columnwidth} 
\caption{CLARIFID Inference-Time Control}
\label{alg:flexnext_core}
\begin{algorithmic}[1]
\REQUIRE Generator $\pi_\theta$, value model $V_\phi$, 
          required \texttt{<next>} token count $k$, 
          number of candidates $N$, 
          temperatures $T_{\mathrm{find}}, T_{\mathrm{imp}}$
\FOR{$j \gets 1$ to $N$}
    \STATE $\mathbf{t} \gets$ ``\texttt{Findings:}''
    \WHILE{$\#\texttt{<next>}(\mathbf{t}) < k$}
        \STATE $y \gets \pi_\theta.\text{sample\_next}(\mathbf{t}, T_{\mathrm{find}})$
        \IF{$y \in \{\texttt{<eos>}, \texttt{<impression>}\}$}
            \STATE $y \gets \texttt{<next>}$
        \ENDIF
        \STATE Append $y$ to $\mathbf{t}$
    \ENDWHILE
    \STATE Append \texttt{<impression>} to $\mathbf{t}$
    \REPEAT
        \STATE $y \gets \pi_\theta.\text{sample\_next}(\mathbf{t}, T_{\mathrm{imp}})$
        \STATE Append $y$ to $\mathbf{t}$
    \UNTIL{$y = \texttt{<eos>}$}
    \STATE $\hat{s}_j \gets \mathbf{t}$, \quad $v_j \gets V_\phi(\hat{s}_j)[-1]$
\ENDFOR
\STATE $j^\ast \gets \arg\max_j v_j$
\RETURN $\hat{s}_{j^\ast}$
\end{algorithmic}
\end{minipage}
\end{algorithm}

\subsection{Inference-time control}
\subsubsection{Next-token forcing}
To ensure sufficiently detailed reports, we adopt a controlled decoding scheme using \texttt{<next>} and \texttt{<impression>} tokens. Impression-related tokens (including \texttt{<impression>} and \texttt{<eos>}) are blocked until the Findings section exceeds a minimum length, preventing early termination and yielding richer, multi-sentence findings. Motivated by budget forcing~\cite{muennighoff2025s1simpletesttimescaling}, we treat the Findings section as intermediate reasoning and the Impression section as the final answer, encouraging grounded and interpretable outputs.

\subsubsection{Best-of-N decoding with temperature scheduling}
\label{BoN impression}
We generate the full Findings–Impression report in a single pass while sampling with distinct temperatures for each section. For the \textbf{Findings}, we use a slightly higher temperature \(T_{\text{find}}\) to encourage diversity and richer detail, helping the model surface subtle abnormalities. For the \textbf{Impression}, we use a lower temperature \(T_{\text{imp}}\) to promote consistency in the clinical summary.

The entire decoding process is repeated \(N\) times, yielding candidate reports.  
Each candidate is assessed by the PPO value model, and the highest-scoring report is chosen.  
This re-ranking step, together with the sentence-count constraint, improves both the Findings and Impression. The complete inference procedure is summarized in algorithm~\ref{alg:flexnext_core}.

\section{Experiments}
\label{sec:experiments}
\subsection{Dataset}
\subsubsection{MIMIC-CXR}
We use the MIMIC-CXR~\cite{johnson2019mimiccxrjpglargepubliclyavailable} dataset, which contains a large collection of chest X-ray images, along with corresponding radiology reports. MIMIC-CXR groups X-ray images and their associated radiology reports by “study.” Each study may contain multiple chest X-ray images, along with a single radiology report. We follow the standard train, validation, and test splits provided in~\cite{johnson2019mimiccxrjpglargepubliclyavailable}. For fair comparison with prior works, our evaluation on the full test set compares only the generated ``Findings'' section with the ground-truth ``Findings'' section.
\subsubsection{MIMIC-CXR Impression}
Under official splitting of MIMIC-CXR, we specifically select only those reports containing both the ``Findings'' and ``Impression'' sections, ensuring that our model is trained on comprehensive reports reflective of real-world clinical workflows. After filtering, we obtain 123,742 training, 977 validation, and 1,600 test studies.
\subsubsection{IU X-ray}
To assess the generalization ability of our approach, we further evaluate it on the public IU X-ray collection~\cite{demner2016preparing}. Following the protocol proposed by~\cite{10.1609/aaai.v38i3.28038}, we train the model exclusively on MIMIC-CXR and evaluate it on the IU X-ray test split defined in that work. For a fair comparison with MIMIC-CXR, we keep only those studies that include both a “Findings” and an “Impression” section, resulting in 4,162 chest-radiograph images paired with 2,081 reports.

\subsection{Evaluation metrics}
Following previous works~\cite{chen-emnlp-2020-r2gen,chen-etal-2021-cross-modal,kiut,COMG}, we measure both clinical efficacy (CE) and natural language generation (NLG) metrics. Specifically, we compute Precision, Recall, and F1-scores (P, R, F1) for CE, and BLEU-1(B-1), BLEU-4(B-4), METEOR(MTR), and ROUGE-L(R-L) for NLG.
For CE, we utilize CheXbert to extract disease labels from the generated text and then calculate micro-averaged P, R, and F1 following~\cite{chen-emnlp-2020-r2gen,chen-etal-2021-cross-modal}. 
NLG metrics are computed using the~\cite{lin2015microsoftcococommonobjects}. In earlier studies that focus only on generating ``Findings'' (without an ``Impression''), evaluation is performed solely on the ``Findings'' section. For models that also produce ``Impression'', we consider both ``Findings'' and ``Impression''.

\subsection{Implementation details}
All components are trained from scratch. We use the AdamW optimizer~\cite{AdamW}. All images are resized to $512\times512$ during training and inference. During section-aware pretraining we optimize the image encoder, the lightweight aggregator, and the GPT-2 decoder end-to-end with a learning rate of $1\times10^{-4}$. To improve robustness we apply light geometric augmentation with random rotation within $\pm5^\circ$. At each epoch we randomly permute the order of Findings sentences, insert \texttt{<next>} between adjacent sentences, and prepend \texttt{<impression>} before the Impression. The policy model uses a 30-layer GPT-2 backbone and the value model uses a 6-layer backbone. We use $C_{\text{pre}}=68$ and $C_{\text{img}}=10$.

In the PPO phase we use a learning rate of $1\times10^{-5}$, a KL divergence penalty coefficient of $0.05$, 4 rollouts per sample, $\epsilon=0.2$, and $\lambda_V=0.1$. The PPO batch size is 4 with gradient accumulation for an effective batch size of 256. Sampling temperature during rollouts is fixed at 1.0.

At inference we use best-of-N with $N=8$. We force up to 10 \texttt{<next>} tokens to maintain sentence boundaries in Findings. Temperatures are $T_{\text{find}}=1.0$ and $T_{\text{imp}}=0.8$ with nucleus sampling $p=0.9$.
\subsubsection{Inference time}
The evaluation was carried out on a single NVIDIA RTX 6000 GPU with a batch size of 2. On average per study, processing the MIMIC-CXR Impression test set took 0.90 s with plain greedy decoding, 6.43 s with next-token forcing to enforce the Findings length, and 13.41 s with both next-token forcing and the BoN strategy.

\begin{table*}[t]
\centering
\caption{Comparison with SOTA methods on MIMIC-CXR (M-CXR), MIMIC-CXR Impression (M-CXR Imp.), and IU X-ray. Methods annotated with the symbol $\dagger$ have been re-evaluated using official codes. The best performance for each dataset is in \textbf{bold}.}
\label{tab:eval_metrics}
\resizebox{\textwidth}{!}{
\begin{tabular}{clcccccccccc}
\toprule
\multirow{2}{*}{\textbf{Dataset}} &
\multirow{2}{*}{\textbf{Method}} &
\multicolumn{3}{c}{\textbf{Findings CE} $\uparrow$} &
\multicolumn{3}{c}{\textbf{Impression CE} $\uparrow$} &
\multicolumn{4}{c}{\textbf{NLG Metrics} $\uparrow$} \\
\cmidrule(lr){3-5} \cmidrule(lr){6-8} \cmidrule(lr){9-12}
 & & P & R & F1 & P & R & F1 & B-1 & B-4 & MTR & R-L \\
\midrule
\multirow{11}{*}{\textbf{M-CXR}} &
R2Gen & 0.333 & 0.273 & 0.276 & -- & -- & -- & 0.353 & 0.103 & 0.142 & 0.277 \\
& R2GenCMN & 0.334 & 0.275 & 0.278 & -- & -- & -- & 0.353 & 0.106 & 0.142 & 0.278 \\
& KiUT & 0.371 & 0.318 & 0.321 & -- & -- & -- & 0.393 & 0.113 & 0.160 & 0.285 \\
& RGRG & 0.461 & 0.475 & 0.447 & -- & -- & -- & 0.373 & 0.126 & 0.168 & 0.264 \\
& METransformer & 0.364 & 0.309 & 0.311 & -- & -- & -- & 0.386 & 0.124 & 0.152 & \textbf{0.362} \\
& R2GenGPT(Deep) & 0.392 & 0.387 & 0.389 & -- & -- & -- & 0.411 & 0.134 & 0.160 & 0.297 \\
& ORGAN & 0.416 & 0.418 & 0.385 & -- & -- & -- & 0.407 & 0.123 & 0.162 & 0.293 \\
& COMG & 0.424 & 0.291 & 0.345 & -- & -- & -- & 0.363 & 0.124 & 0.128 & 0.290 \\
& EKAGen & 0.517 & 0.483 & 0.499 & -- & -- & -- & \textbf{0.419} & 0.119 & 0.157 & 0.287 \\
& MLRG & \textbf{0.549} & 0.468 & 0.505 & -- & -- & -- & 0.411 & \textbf{0.158} & \textbf{0.176} & 0.320 \\
& Ours & 0.514 & \textbf{0.584} & \textbf{0.547} & -- & -- & -- & 0.273 & 0.057 & 0.167 & 0.174 \\
\midrule
\multirow{5}{*}{\shortstack[c]{\textbf{M-CXR}\\\textbf{Imp.}}} &
R2Gen$\dagger$ & 0.379 & 0.196 & 0.258 & -- & -- & -- & 0.378 & 0.113 & 0.145 & 0.282 \\
& R2GenCMN$\dagger$ & 0.413 & 0.298 & 0.347 & -- & -- & -- & 0.361 & 0.108 & 0.144 & 0.286 \\
& PromptMRG$\dagger$ & \textbf{0.564} & 0.484 & 0.521 & -- & -- & -- & 0.397 & 0.106 & 0.150 & 0.267 \\
& R2GenGPT(Deep)$\dagger$ & 0.400 & 0.327 & 0.360 & 0.398 & 0.362 & 0.379 & \textbf{0.406} & \textbf{0.135} & 0.114 & \textbf{0.305} \\
& Ours & 0.506 & \textbf{0.557} & \textbf{0.530} & \textbf{0.458} & \textbf{0.540} & \textbf{0.496} & 0.325 & 0.073 & \textbf{0.179} & 0.207 \\
\midrule
\multirow{5}{*}{\shortstack[c]{\textbf{IU}\\\textbf{X-ray}}} &
R2Gen$\dagger$ & 0.176 & 0.181 & 0.179 & -- & -- & -- & 0.292 & 0.054 & 0.140 & 0.255 \\
& R2GenCMN$\dagger$ & 0.180 & 0.242 & 0.207 & -- & -- & -- & 0.387 & 0.085 & 0.146 & 0.283 \\
& PromptMRG$\dagger$ & \textbf{0.306} & \textbf{0.343} & \textbf{0.323} & -- & -- & -- & \textbf{0.409} & \textbf{0.105} & 0.152 & 0.306 \\
& R2GenGPT(Deep)$\dagger$ & 0.236 & 0.251 & 0.244 & 0.192 & 0.234 & 0.211 & 0.405 & 0.093 & 0.123 & \textbf{0.325} \\
& Ours & 0.277 & 0.281 & 0.279 & \textbf{0.611} & \textbf{0.667} & \textbf{0.638} & 0.243 & 0.050 & \textbf{0.195} & 0.218 \\
\bottomrule
\end{tabular}
}
\end{table*}

\begin{table*}[t]
\centering
\caption{Results of ablation experiments on MIMIC-CXR Imp. dataset. The best results are in \textbf{bold}.}
\label{tab:ablation_mimic}
\resizebox{\columnwidth}{!}{
\begin{tabular}{cccccc|ccc|ccc|cccc}
\toprule
\multicolumn{6}{c|}{\textbf{Method}} & \multicolumn{3}{c|}{\textbf{Findings CE} $\uparrow$} & \multicolumn{3}{c|}{\textbf{Impression CE} $\uparrow$} & \multicolumn{4}{c}{\textbf{NLG Metrics} $\uparrow$} \\
\cmidrule(lr){1-6} \cmidrule(lr){7-9} \cmidrule(lr){10-12} \cmidrule(lr){13-16}
SL & MV & SA & RL & FG & BoN & P & R & F1 & P & R & F1 & B-1 & B-4 & MTR & R-L \\
\midrule
\cmark &  & \cmark &  &  &  & 0.468 & 0.378 & 0.419 & 0.363 & 0.317 & 0.338 & 0.274 & 0.059 & 0.154 & 0.204 \\
\cmark & \cmark &  &  &  &  & 0.499 & 0.406 & 0.448 & 0.403 & 0.360 & 0.380 & 0.217 & 0.048 & 0.137 & 0.195 \\
\cmark & \cmark & \cmark &  &  &  & 0.532 & 0.460 & 0.493 & 0.437 & 0.397 & 0.416 & 0.276 & 0.062 & 0.157 & 0.206 \\
\cmark & \cmark & \cmark & \cmark &  &  & \textbf{0.546} & 0.474 & 0.507 & \textbf{0.477} & 0.482 & 0.479 & 0.234 & 0.053 & 0.148 & 0.206 \\
\cmark & \cmark & \cmark & \cmark & \cmark &  & 0.521 & 0.529 & 0.525 & 0.459 & 0.519 & 0.487 & \textbf{0.329} & \textbf{0.075} & 0.178 & 0.207 \\
\cmark & \cmark & \cmark & \cmark & \cmark & \cmark & 0.506 & \textbf{0.557} & \textbf{0.530} & 0.458 & \textbf{0.540} & \textbf{0.496} & 0.325 & 0.073 & \textbf{0.179} & \textbf{0.207} \\
\bottomrule
\end{tabular}
}
\end{table*}

\section{Results}
\subsection{Comparison with SOTA}
Table~\ref{tab:eval_metrics} contrasts our model with recent state-of-the-art systems; R2Gen~\cite{chen-emnlp-2020-r2gen}, R2GenCMN~\cite{chen-etal-2021-cross-modal}, KiUT~\cite{kiut}, RGRG~\cite{RGRG}, METransformer~\cite{METransformer}, 
R2GenGPT~\cite{wang2023r2gengptradiologyreportgeneration},ORGAN~\cite{gu2023complexorganmaskguided}, COMG~\cite{COMG}, EKAGen~\cite{EKAGen}, MLRG~\cite{MLRG}, PromptMRG~\cite{10.1609/aaai.v38i3.28038}.
On the MIMIC-CXR Impression test set we outperform all baselines, including those optimized solely for the Findings section. Our method attains the best Impression CE (P 0.458 / R 0.540 / F1 0.496) and the highest Findings CE recall (0.557) and F1 (0.530), evidencing superior recovery of clinically pertinent details. {Trained only on MIMIC-CXR, our model also leads on IU X-ray (Impression-CE F1 0.638) without domain adaptation. In this split, 39.1\% of Findings contain no positive CheXbert labels and 60.6\% of Impressions are “No Findings.” Because of this asymmetry, Findings—where labels are scarce—tend to yield more restrained scores, while Impression is heavily influenced by the correct identification of ``No Finding,'' which dominates its label distribution.} Together, these gains underline the clinical reliability and robustness of our joint Findings-and-Impression optimization strategy. {Note that MIMIC-CXR serves as a Findings-only benchmark, so a broader set of methods is included. In contrast, MIMIC-CXR Impression and IU X-ray require an explicit Impression for section-aware evaluation, and only a subset of baselines provide such outputs.}

\subsection{Ablation study}
Table~\ref{tab:ablation_mimic} evaluates the impact of each component on MIMIC-CXR Impression.
Turning on multi-view (+MV, using all radiographs within a study) strengthens the supervised baseline, improving ``Findings'' F1 from 0.419 to 0.493 and ``Impression'' F1 from 0.338 to 0.416, indicating that complementary views provide stronger evidence.
{Under the same MV configuration, removing section-aware supervision (SA) degrades both CE and NLG: ``Findings'' F1 0.493$\to$0.448, ``Impression'' F1 0.416$\to$0.380; BLEU-1 0.276$\to$0.217, BLEU-4 0.062$\to$0.048, METEOR 0.157$\to$0.137, ROUGE-L 0.206$\to$0.195, suggesting that section-aware supervision provides helpful structure for organizing evidence and summarization. The ablation study for MV and SA was conducted in the same training epochs.}
Adding reinforcement learning (+RL) that maximizes ``Impression'' F1 uplifts both ``Findings'' and ``Impression'' scores, showing that summary-level rewards inject clinically useful signals into the entire reasoning pipeline.  
Introducing the forced generation (+FG) raises F1 for both sections, supporting our key claim that radiology report generation is inherently sequential—accurate ``Findings'' naturally set the stage for concise ``Impression''.  
Finally, the Best-of-N (+BoN) provides the strongest overall performance, raising ``Findings'' F1 to 0.530, ``Impression'' F1 to 0.496, and METEOR to 0.179.

\begin{figure}[t]
  \centering
  \includegraphics[width=0.5\columnwidth]{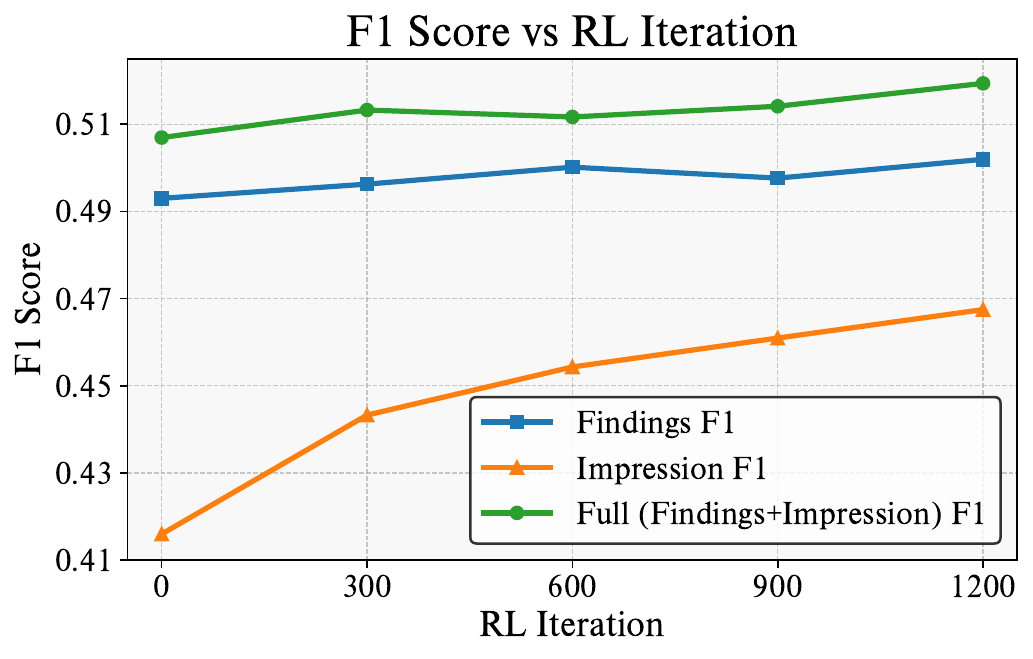}
  \caption{Section-wise {and overall F1} scores on the MIMIC-CXR Impression through RL iteration.}
  \label{fig:effect_of_rl}
\end{figure}

\subsection{Section-wise effects of reinforcement learning}
To obtain a more detailed analysis of the effect of reinforcement learning (RL), we tracked section-level F1 score during reinforcement learning training. Figure \ref{fig:effect_of_rl} presents these scores on the MIMIC-CXR Impression test set across RL iterations. The Impression F1 increases steadily before saturating at approximately $0.465$, whereas the Findings F1 rises and plateaus near $0.5$ over the same interval. These parallel improvements indicate that an Impression-based reward not only strengthens the Impression section itself but also yields measurable gains in the Findings section.

{\subsection{Intra-report consistency of Findings and Impression}

\begin{table}[t]
\centering
\caption{CheXbert intra-report consistency on MIMIC-CXR Impression.}
\setlength{\tabcolsep}{6pt}
\begin{tabular}{lcc}
\toprule
Method & $R_e$ & $P_e$\\
\midrule
Ground Truth & 0.569 & 0.413\\
R2GenGPT(Deep) & 0.189 & 0.160\\
Ours(SL) & 0.508 & 0.367\\
Ours(SL + RL) & 0.616 & 0.469\\
\bottomrule
\end{tabular}
\label{tab:chexbert_consistency_mimic_imp}
\end{table}

\begin{table}[t]
\centering
\caption{CheXbert intra-report consistency on IU X-ray.}
\setlength{\tabcolsep}{6pt}
\begin{tabular}{lcc}
\toprule
Method & $R_e$ & $P_e$\\
\midrule
Ground Truth & {0.224} & {0.211}\\
R2GenGPT(Deep) & 0.038 & 0.027\\
Ours & {0.159} & {0.132}\\
\bottomrule
\end{tabular}
\label{tab:chexbert_consistency_iux}
\end{table}

To verify that the diagnostic Impression is supported by the evidence described in the Findings, we introduce a CheXbert-based intra-report consistency probe that measures alignment between disease-level positives stated in the Impression and those documented in the Findings. Prior work has examined this axis—treating discrepancies between the Findings and the Impression sections as detectable errors~\cite{RRED} and incorporating ``impression consistency'' into a radiologist-derived evaluation framework~\cite{MRScore}. However, to preserve direct comparability with standard Clinical Efficacy (CE) reporting~\cite{chen-emnlp-2020-r2gen}, we introduce two summary metrics—evidence recall and evidence precision. This direct comparability to CE allows improvements to be interpreted on the same label set and scale widely used across datasets and baselines.\\
For study condition $c\in\{1,\dots,13\}$ (13 disease labels) and section $s\in\{F, I\}$ (Findings, Impression). We get binarized CheXbert label as in Eq.~\eqref{eq:binary_label}:
\[
g_{s,c} = \mathbbm{1}[\tilde{g}_{s,c} = \text{positive}] \in \{0, 1\}
\]
Let $L^+_{F}=\{c:g_{F,c}=1\}$ and $L^+_{I}=\{c:g_{I,c}=1\}$ denote the sets of positive labels from Findings and Impression sections of report, respectively. Evidence recall ($R_e$) measures the fraction of positive labels asserted in the Impression that are present as evidence in the Findings, whereas evidence precision ($P_e$) measure the fraction of positive labels enumerated in Findings that are carried forward into the Impression.
\[
R_e = \frac{|{L}_{I}^+ \cap {L}_{F}^+|}{\max(1, |{L}_{I}^+|)}, \quad
P_e = \frac{|{L}_{I}^+ \cap {L}_{F}^+|}{\max(1, |{L}_{F}^+|)}.
\]
Since the goal of $(R_e, P_e)$ is to test whether positive diagnostic claims in the Impression are evidenced in the Findings, we compute them over the 13 disease labels, excluding CheXbert's ``No Findings'' label.\\
Table~\ref{tab:chexbert_consistency_mimic_imp} shows a clear improvement obtained by incorporating reinforcement learning (RL) on MIMIC-CXR Impression (
\(
R_e: 0.508\rightarrow0.616,\; P_e: 0.367\rightarrow0.469
\)
). These results reinforce that optimizing Impression-level clinical correctness not only improves section-wise CE but also tightens the coupling between sections: a larger fraction of Impression positives are explicitly present in the Findings, and the Findings content reflected in the Impression becomes more focused.
Moreover, compared to R2GenGPT~\cite{wang2023r2gengptradiologyreportgeneration}, our supervised baseline (SL) already achieves markedly higher consistency ($\Delta R_e{=}{+}0.319$, $\Delta P_e{=}{+}0.207$). This pattern suggests that our section-aware training also enhances the linkage between the two sections.
On IU X-ray (Table~\ref{tab:chexbert_consistency_iux}), our model—trained only on MIMIC-CXR outperforms R2GenGPT by a wide margin ($R_e$: 0.159 vs.\ 0.038, $P_e$: 0.132 vs.\ 0.027), supporting our claim of better cross-dataset consistency. The remaining gap to ground-truth levels reflects domain/style shift.
}

\subsection{Effects of next-token forcing}
\begin{figure}[t]
    \centering
    \includegraphics[width=0.5\columnwidth]{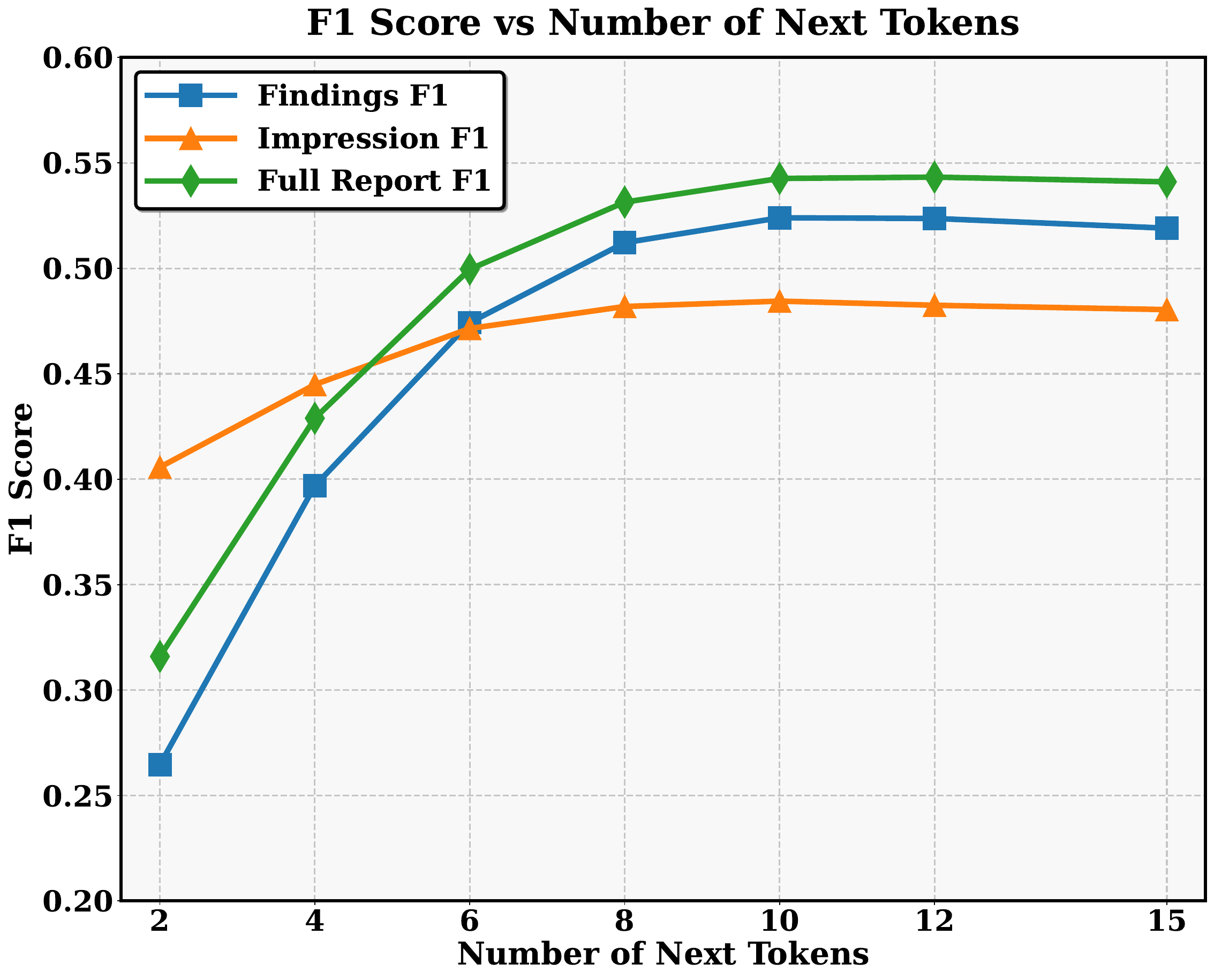}
    \caption{Impact of next-token forcing count on clinical accuracy F1 scores for Findings, Impression, and Full Report. Scores are plotted across forcing lengths of 2 to 15 tokens without BoN.}
    \label{fig:effect_of_next_token}
\end{figure}

We measured clinical efficacy (CheXbert) F1 scores on both the Impression and Findings sections while varying the number of forced “next” tokens during decoding ($n = 2, 4, 6, 8, 10, 12, 15$). As shown in Figure~\ref{fig:effect_of_next_token}, increasing the next-token budget consistently improves performance up to $n=10$, after which further forcing yields diminishing returns. At the lower end ($n=2$), the model struggles to recover many disease labels, reaching only $0.406$ F1 on Impression and $0.265$ on Findings. The largest gains occur between $n=2$ and $n=6$ ($\Delta{+}0.065$ for Impression and $\Delta{+}0.209$ for Findings), as increasing the next-token budget transforms a report that previously lacked room to convey all pertinent findings into one that can accommodate substantially more clinical information. The highest micro-F1 scores are achieved at $n=10$ ($0.487$ for Impression, $0.525$ for Findings), while further increases beyond $n=10$ result in marginal or no improvement.

\begin{table}[t]
  \centering
\caption{Performance at different temperature scheduling settings.}
  \setlength{\tabcolsep}{1mm}
  \begin{tabular}{lcccc}
    \toprule
    \textbf{Metric} &
    \begin{tabular}[c]{@{}c@{}}$T_{\text{f}}=1.2$\\$T_{\text{i}}=0.8$\end{tabular} &
    \begin{tabular}[c]{@{}c@{}}$T_{\text{f}}=1.0$\\$T_{\text{i}}=1.0$\end{tabular} &
    \begin{tabular}[c]{@{}c@{}}$T_{\text{f}}=1.0$\\$T_{\text{i}}=0.8$\end{tabular} &
    \begin{tabular}[c]{@{}c@{}}$T_{\text{f}}=1.0$\\$T_{\text{i}}=0.5$\end{tabular} \\
    \midrule
    Find. P      & 0.498 & \textbf{0.514} & 0.506 & 0.502 \\
    Find. R      & \textbf{0.562} & 0.560 & 0.557 & 0.550 \\
    Find. F1     & 0.528 & \textbf{0.536} & 0.530 & 0.525 \\
    \midrule
    Imp. P    & 0.444 & 0.453 & \textbf{0.458} & 0.445 \\
    Imp. R    & 0.526 & 0.537 & \textbf{0.540} & 0.527 \\
    Imp. F1   & 0.481 & 0.491 & \textbf{0.496} & 0.482 \\
    \midrule
    Full. B‑1 & 0.321 & \textbf{0.327} & 0.325 & 0.324 \\
    Full. B‑4 & 0.070 & \textbf{0.074} & 0.073 & 0.072 \\
    Full. MTR & 0.179 & \textbf{0.180} & 0.179 & 0.179 \\
    Full. R‑L & 0.204 & 0.205 & \textbf{0.206} & \textbf{0.206} \\
    \bottomrule
  \end{tabular}
  \label{tab:temperature_results}
\end{table}

\subsection{Effects of temperature scheduling}
\label{sec:temp_sched}
To identify an effective division of ``creativity’’ Findings and ``conciseness’’ Impression at inference time, we conduct a small grid‑search over the sampling temperatures \(T_{\text{find}}(T_{\text{f}})\) and \(T_{\text{imp}}(T_{\text{i}})\).
We explored the four settings as follows.
\[(T_{\text{find}},T_{\text{imp}})\in\!\{(1.2,0.8),(1.0,1.0),
(1.0,0.8),(1.0,0.5)\}\]
Table~\ref{tab:temperature_results} summarizes the outcome. The balanced setting \((1.0,1.0)\) yields the best Findings \(\text{F1}=0.536\), whereas the hotter Findings decoder \((1.2,0.8)\) sacrifices precision for recall, resulting in a slightly lower F1. Cooling the Impression decoder to \(T_{\text{imp}}=0.8\) while keeping \(T_{\text{find}}=1.0\) gives the highest Impression \(\text{F1}=0.496\), confirming that lower temperature helps the model compose a clearer clinical summary.

\begin{figure*}[t]
	\centering
	\includegraphics[width=0.7\textwidth]{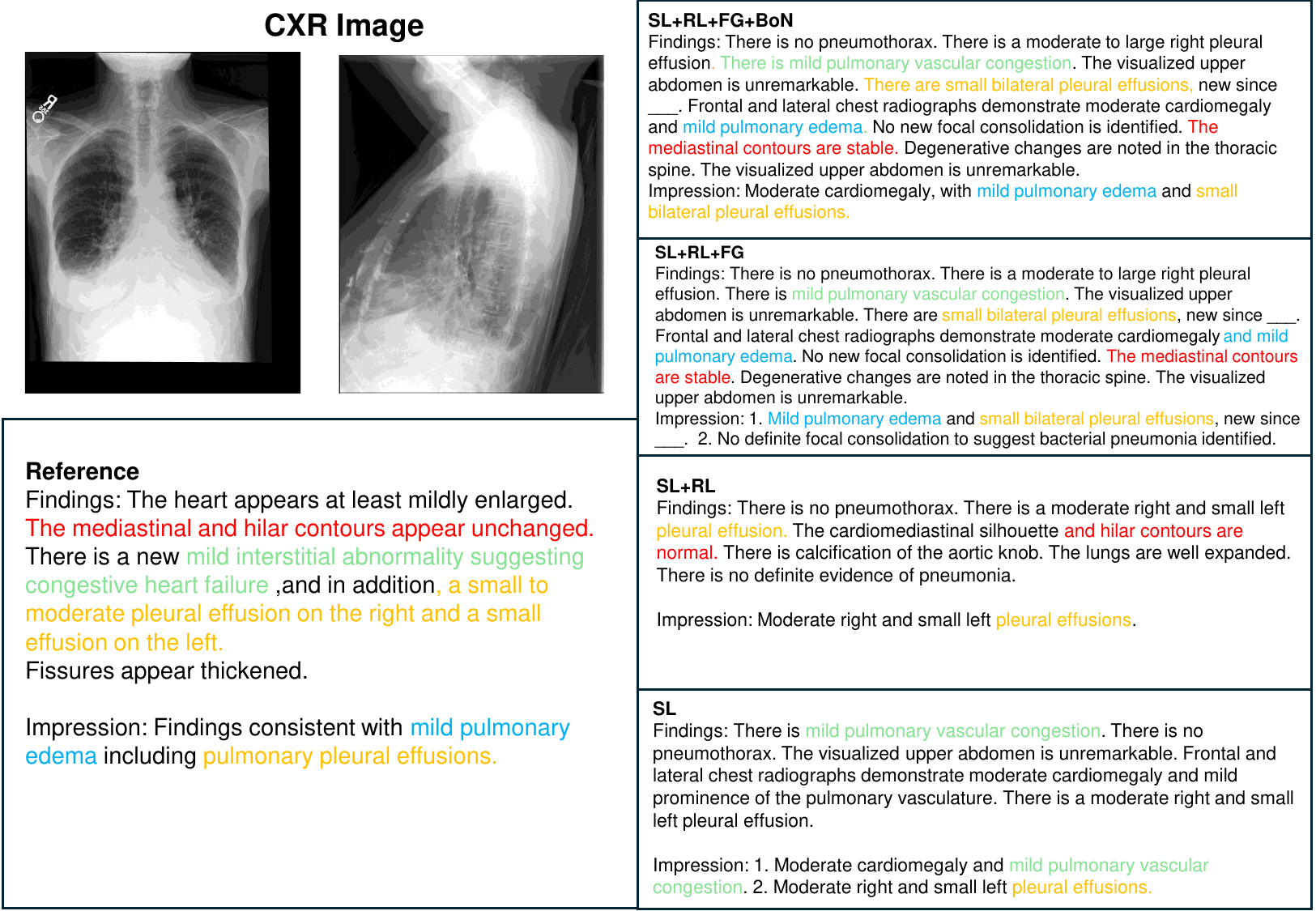}
	\caption{Qualitative ablation study of MIMIC-CXR report generation: semantic-level color highlighting of outputs under conditions versus the reference report.
	}
	\label{fig:test_case}
\end{figure*}

\begin{figure*}[!t]
	\centering
	\includegraphics[width=0.7\textwidth]{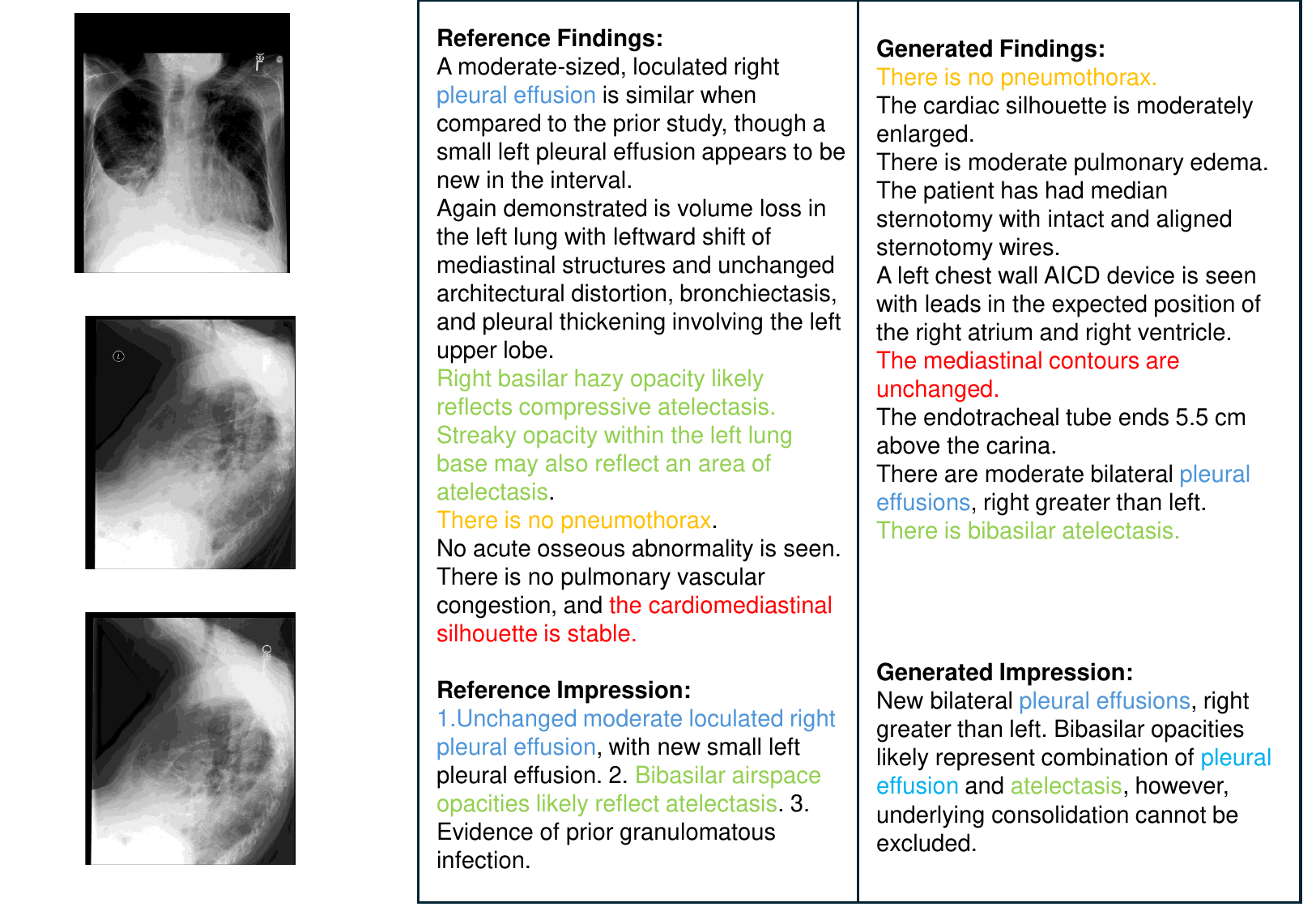}
	\caption{Additional MIMIC-CXR dataset test case}
	\label{fig:test_case2}
\end{figure*}

\begin{figure*}[!t]
	\centering
	\includegraphics[width=0.7\linewidth]{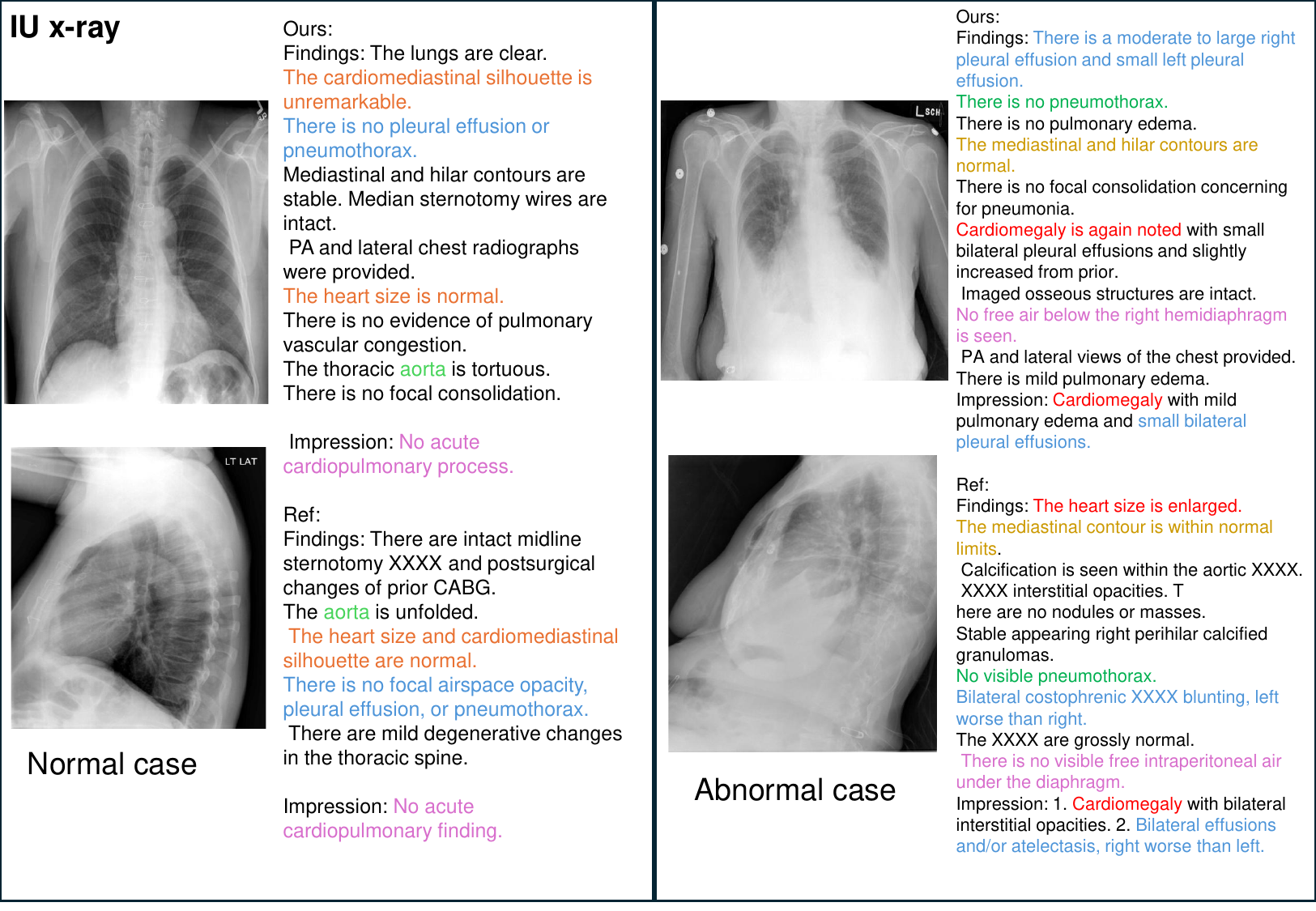}
	\caption{Qualitative evaluation on IU X-ray,
    The term ‘xxxx’ in the reference report denotes keywords that were removed during the de-identification process }
	\label{fig:iu_xray_qual}
\end{figure*}

{\subsection{Classification head performance}
\label{sec:cls_performance}

To quantify the disease-prediction component used during section-aware pretraining, we evaluate the 14-label classification head on a held-out split. As summarized in Table \ref{tab:classifier_micro}, the model attains micro-averaged Precision = 0.438,Recall = 0.450, and F1 = 0.419. These results indicate that the visual encoder captures clinically informative cues. note that this  classification head is used for pretraining/analysis and does not participate in report generation at inference time.
}

\begin{table}
\centering
\caption{{Multi-label chest X-ray classification performance of the visual encoder.}}
\small
\begin{tabular}{lccc}
\toprule
Average & Precision & Recall & F1 \\
\midrule
Micro   & 0.438     & 0.450  & \textbf{0.419} \\
\bottomrule
\end{tabular}
\label{tab:classifier_micro}
\end{table}

\subsection{Qualitative results}
To demonstrate the clinical faithfulness and generalizability of our model, we conducted a qualitative evaluation on both the MIMIC-CXR and IU X-ray datasets. For our analysis, we overlaid color highlights on text segments where the generated report semantically matched the ground-truth reference, allowing for a direct comparison of clinical content.
\subsubsection{MIMIC-CXR results}
On the MIMIC-CXR test set, our model consistently generates semantically rich and clinically accurate reports. As shown in Figures~\ref{fig:test_case} and \ref{fig:test_case2}, our controlled budget-forcing strategy demonstrates clear benefits; increasing the decoding budget at test time leads to a higher degree of overlap with the reference, yielding more clinically faithful reports. Furthermore, the model successfully preserves key findings (e.g., thoracic stability) and ensures that critical diagnostic information identified in the Findings section is coherently reiterated in the Impression section. This semantic integrity is maintained even when the generated report uses different phrasing than the ground-truth reference.

\begin{figure}
    \centering
    \includegraphics[width=0.7\linewidth]{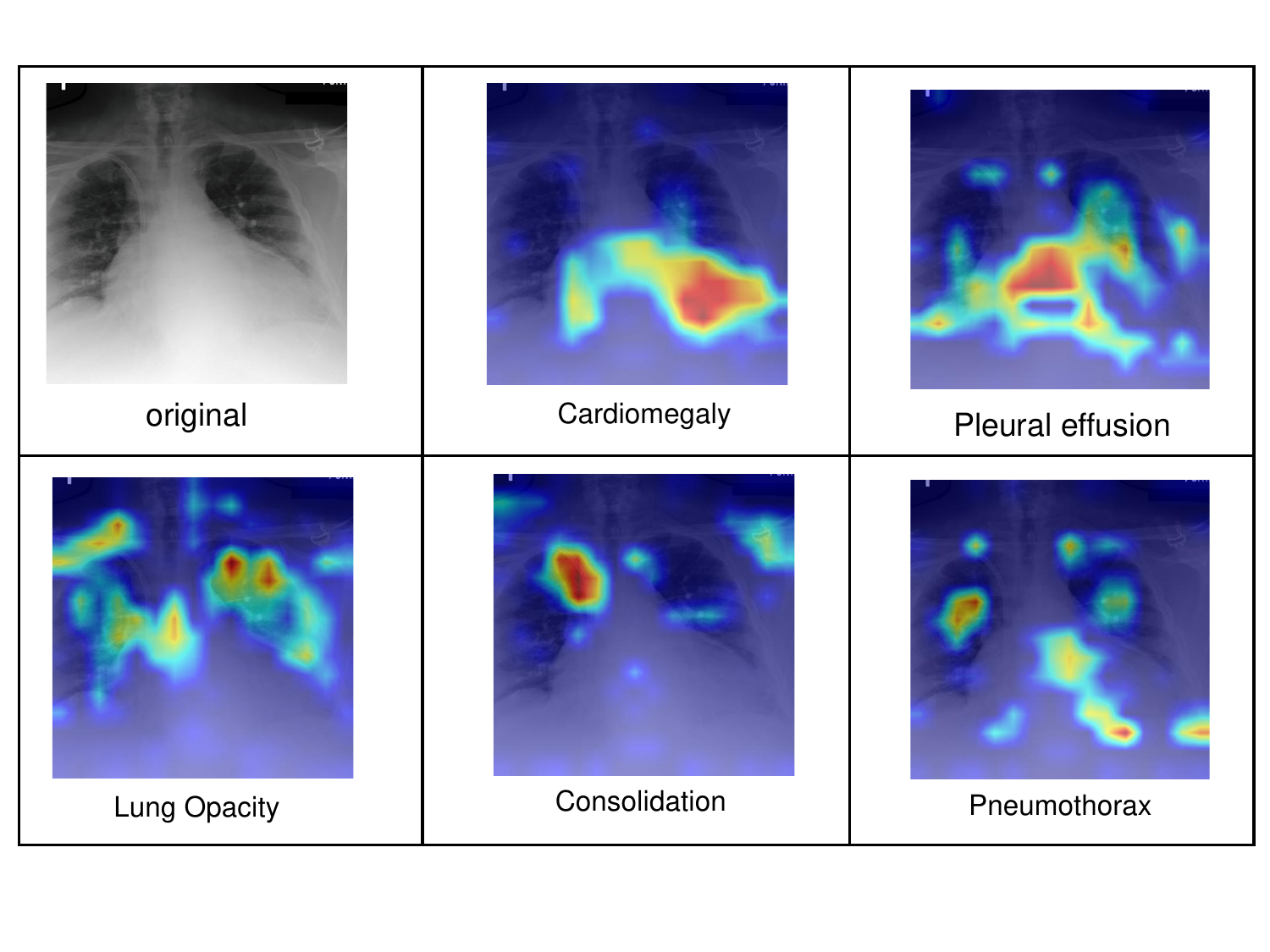}
    \caption{{Class-targeted Grad-CAM visualizes organ/region-specific heatmaps used by the model when composing the report.}}
    \label{fig:grad_cam}
\end{figure}

{As shown in figure~\ref{fig:grad_cam} class-targeted Grad-CAM ~\cite{Selvaraju_2019} visualizations indicate that the model’s attention is appropriately concentrated on the relevant anatomical regions associated with each diagnostic concept. This suggests that the model does not rely on spurious correlations, but instead focuses on clinically meaningful image regions during the reasoning process that precedes report generation. Overall, the qualitative heatmaps confirm that the visual encoder successfully captures disease-related cues from the appropriate anatomical structures.}

\subsubsection{IU X-ray results}
To assess the model's performance in a different domain without specific fine-tuning, we evaluated it on the IU X-ray dataset. The representative examples in Figure~\ref{fig:iu_xray_qual} showcase the model's strong generalizability. It accurately identifies key abnormal findings (e.g., pleural effusion, cardiomegaly) in pathological cases while avoiding hallucinated findings in normal cases, appropriately reflecting the absence of significant abnormalities. These results confirm that our model maintains high clinical consistency and interpretability across both typical and challenging scenarios, even on unseen data.

\section{Conclusion}
We have presented \textbf{CLARIFID}, a unified radiology-report generation framework that explicitly mirrors the radiologist’s two-step reasoning process—first documenting ``Findings'', then synthesizing an ``Impression''.  By directly optimizing the clinical correctness of the Impression section and employing controlled decoding, our approach aligns both training and inference with real-world diagnostic practice. This synergy delivers state-of-the-art performance across benchmarks, boosting Impression-level clinical efficacy on MIMIC-CXR and, without any domain tailoring, achieves the best Impression-level F1 on IU X-ray, providing strong evidence of robust cross-dataset generalization.

\section*{Limitations}
Although CLARIFID markedly improves clinical fidelity, several limitations remain. First, prioritizing impression-level accuracy can narrow linguistic diversity, yielding lower BLEU and ROUGE scores than text-centric baselines. Second, our evaluation relies entirely on automatic metrics; blinded reviews by practicing radiologists are still needed to verify real-world utility and expose subtle errors. Future work will seek a better balance between clinical faithfulness and language naturalness, and incorporate expert human assessment to address these issues.
\section*{CRediT authorship contribution statement}

\textbf{Kyeongkyu Lee}: Writing—original draft, Writing—review \& editing, Visualization, Validation, Investigation, Conceptualization, Data curation, Software.  

\textbf{Seonghwan Yoon}: Writing—original draft, Writing—review \& editing, Visualization, Validation, Investigation, Conceptualization, Data curation, Software. 

\textbf{Hongki Lim}: Writing—review \& editing, Supervision, Conceptualization, Methodology, Project administration, Data curation, Software, Resources.

\section*{Declaration of generative AI and AI-assisted technologies in the writing process}
During the preparation of this work the authors used ChatGPT in order to improve the clarity and readability of the manuscript. After using this service, the authors reviewed and edited the content as needed and take full responsibility for the content of the published article.

\section*{Declaration of Competing Interest}
The authors declare that they have no known competing financial interests or personal relationships that could have appeared to influence the work reported in this paper.

\section*{Acknowledgment}
This work was supported in part by the National Research Foundation of Korea (NRF) grant funded by the Korea government (MSIT) (RS-2025-24683103), in part by Korea Basic Science Institute (National Research Facilities and Equipment Center) grant funded by the Ministry of Science and ICT (No. RS-2024-00401899), and in part by Institute of Information \& Communications Technology Planning \& Evaluation (IITP) under the Leading Generative AI Human Resources Development (IITP-2025-RS-2024-00360227) grant funded by the Korea government (MSIT).

\section*{Data Availability}
All data used in this work are from publicly available sources.

\clearpage
\bibliographystyle{cas-model2-names}
\bibliography{cas-refs}

\end{document}